\documentclass{article}

\usepackage[english]{babel}
\usepackage{comment}

\usepackage[letterpaper,top=2cm,bottom=2cm,left=3cm,right=3cm,marginparwidth=1.75cm]{geometry}

\usepackage[dvipsnames]{xcolor}
\usepackage{multirow}
\usepackage{amsmath}
\usepackage{graphicx}
\usepackage[colorlinks=true, allcolors=blue]{hyperref}

\title{The Role of Rigor in Artificial Intelligence}

\author{Timothy Nguyen\footnote{The author thanks Marcus Hutter, Matthew McAteer, and Tim Scarfe for their invaluable comments and suggestions.}\\[.5ex]
 Google DeepMind\\[.5ex]
  \texttt{timothycnguyen@google.com}}
\date{}

\begin{document}
\maketitle

\begin{abstract}
Artificial intelligence (AI) has achieved extraordinary capabilities despite lacking many of the conceptual and scientific foundations associated with mature disciplines.
Unlike traditional sciences, where reliable technology typically emerges from theoretical understanding, modern AI has progressed largely through performance-driven iteration and ``alchemical'' experimentation.
This tension motivates a systematic analysis of AI through the lens of rigor.
We introduce a three-part framework consisting of conceptual rigor (clarifying foundational concepts), epistemic rigor (establishing scientific understanding), and operational rigor (ensuring reliable performance and deployment).
Using this framework, we analyze competing conceptions of intelligence and understanding, the strengths and limitations of the empirical approach to deep learning, the power and pitfalls of benchmarks, and the obstacles to theory development posed by modern AI systems.
We argue that the distinctive trajectory of AI arises from how forms of rigor interact across paradigms, resulting in the primacy of operational rigor in modern deep learning.
This perspective helps explain both AI's rapid advances and its persistent uncertainties, while clarifying the challenges involved in transforming AI into a mature science and reliable technology.
\end{abstract}

Rigorous science has enabled humanity to understand the world and build reliable technology.
By formulating testable hypotheses and developing increasingly general theories, the sciences have progressively reduced uncertainty while broadening explanatory power.
These advances have yielded technologies whose reliability derives not merely from empirical success but from principled understanding.
Physics predicts the path of rockets, chemistry models molecular interactions underlying drug design, and evolutionary biology explains the diversity of species through natural selection.

Modern artificial intelligence (AI) departs sharply from this pattern.
Current AI systems exhibit remarkable capabilities despite limited understanding of how or why they work.
Progress is often driven less by explanatory theory than by large-scale experimentation and performance on benchmarks.
At the same time, many of the field’s core terms, such as intelligence and understanding, remain ambiguous and contested.
Together, these disparities raise questions about how scientific and conceptual progress in AI should be evaluated.
This motivates examining AI from the standpoint of rigor.
Here, rigor denotes the disciplined use of methods and standards that support the aims of a field.
This includes axiomatic-deductive proof in mathematics and stringent experimental testing in the natural sciences.
A natural difficulty in formulating rigor in AI is that the field is multidisciplinary: it spans domains such as computer science, statistics, engineering, cognitive science, neuroscience, and philosophy, each governed by different standards and aims.
Consequently, rigor cannot be treated as a single, uniform standard, but must instead be understood in accordance with the different functions it serves.
Distinguishing these forms of rigor provides a coherent framework for analyzing the diverse problems and tensions within AI.
It also enables an understanding of the distinctive structure of progress and paradigms throughout the history of AI.

We propose a three-part division for rigor in AI: conceptual, epistemic, and operational.\footnote{An alternative analysis of rigor in AI is offered in \cite{olteanu2025rigor}, which proposes a six-part division oriented toward responsible AI research practices.}
Conceptual rigor concerns the clarity of foundational concepts and paradigms.
Epistemic rigor concerns the establishment of scientific knowledge and understanding.
Operational rigor concerns the reliable deployment and performance of systems in practice.
Together, these forms of rigor are sufficiently broad to incorporate many of the salient problems and ambitions within AI.
But more than merely providing a descriptive framework, they interact across paradigms in ways that have shaped both the structure of progress in AI and the predominance of performance-driven development in modern deep learning.
By identifying the resulting gaps and uneven development, we clarify where further rigor is needed for AI to mature as a scientific and technological discipline.

\section{Conceptual Rigor}
\label{sec:ConceptualRigor}

Conceptual rigor involves the formulation of clear and consistent terminology.
This includes the use of exact definitions, substantive descriptions, or illustrative examples to elucidate core concepts.
Such precision enables proper usage of terms, a prerequisite for subsequent development.
Additionally, conceptual rigor demands the clear articulation of paradigms: foundational frameworks that introduce axiomatic assumptions and basic methodologies.
Paradigms integrate terms, axioms, and methods into a coherent framework for formulating and evaluating claims.

The development of AI has been shaped by such conceptual frameworks from its earliest beginnings.
One of its foundational ideas was the notion of general-purpose computation.
This notion was first conceived by Charles Babbage and Ada Lovelace in the 19th century through their analysis of the Analytical Engine, the earliest attempt at a digital computer.
A century later, Alan Turing formally characterized universal computation through the universal Turing machine, and subsequently suggested that machines might learn rather than rely entirely on explicitly programmed rules, foreshadowing modern machine learning \cite{Turing1950Computing}.
Central to these developments was the idea that a single machine could in principle perform any computable operation given an appropriate encoding.

Drawing from these ideas, the term ``artificial intelligence'' was coined by John McCarthy in a 1956 workshop proposal grounded in ``the conjecture that every aspect of learning or any other feature of intelligence can in principle be so precisely described that a machine can be made to simulate it'' \cite{McCarthy1955Dartmouth}.
The proposal further envisioned machines that could use language, form abstractions, solve problems, and improve themselves.
Given the proposal’s pragmatic aim to secure funding, it made no attempt to define (artificial) intelligence\footnote{The name artificial intelligence was conceived as a practical alternative to other terms such as automata studies and complex information processing \cite{FloridiNobre2024}.}.
In the absence of a standard definition, early AI research proceeded by operationalizing intelligence in terms of specific capabilities, such as reasoning, problem-solving, and language use, each grounded in distinct theoretical frameworks.
This lack of a shared conception meant that judgments of progress were often internal to particular research programs, and over time, what counted as ``intelligence'' shifted as systems succeeded on previously challenging tasks.

The absence of conceptual agreement about intelligence is not merely a semantic inconvenience.
In scientific practice, concepts play a normative role: they determine what phenomena are to be explained, what counts as evidence, and how progress is to be measured.
Without a sufficiently clear characterization of intelligence, evaluations of AI systems risk becoming fragmented across tasks and research programs, thereby undermining the possibility of cumulative progress.
Achieving conceptual rigor in AI thus requires a clear notion of intelligence: more than just a prerequisite for effective communication, this foundational clarity is essential for the coherence of the field itself.

Over the years, many definitions have been proposed for intelligence \cite{wang1995non_axiomatic_reasoning, LeggHutter2007Definitions, Chollet2019MeasureIntelligenceARC}.
In \cite{LeggHutter2007Definitions}, a collection of 70-odd definitions is provided for comparison, as obtained from dictionaries, encyclopedias, psychologists, and AI researchers.
While there is much in common among these definitions, it is perhaps unrealistic to expect that a universal one could be obtained with enough effort and synthesis.
Given this, it is more reasonable to provide not a definition but instead a characterization of intelligence that draws upon its commonly acknowledged features:\\

\noindent\textit{Intelligence is a multifaceted property arising from many factors that each vary across a spectrum, including
\begin{itemize}
\item Learning and problem-solving
\item Adaptability and skill acquisition
\item Goal achievement, planning, and prediction
\item Knowledge and understanding
\item Reasoning and abstraction
\item Efficiency under resource constraints.
\end{itemize}
}

Viewed through this characterization, intelligence is not binary but varies across multiple dimensions.
This diversity helps explain conflicting views on frontier AI systems: judgments differ depending on whether one emphasizes their achievements or their limitations.
In \cite{bubeck2023sparksartificialgeneralintelligence}, the authors perform an extensive investigation of the GPT-4 large language model (LLM), concluding that ``Given the breadth and depth of GPT-4’s capabilities, we believe that it could reasonably be viewed as an early (yet still incomplete) version of an artificial general intelligence (AGI) system.''
More strongly, Geoffrey Hinton believes that today’s large language models are intelligent, and moreover understand language and can have experiences in the same way humans do \cite{60MinutesHinton2023, RoyalInstitutionHinton2025}.\footnote{Hinton does not appear to offer an explicit definition of intelligence or understanding.}
On the other hand, Yann LeCun has declared that current AI systems are not even as smart as a household cat, noting how AI systems cannot plan or reason well and do not understand the world \cite{WorldGovernmentsSummitLeCun2024}.
In \cite{krakauer2025largelanguagemodelsemergence}, which emphasizes among other things the efficiency dimension of intelligence, it is noted that ``There is little reason to expect LLMs to be intelligent since all we have been training through endless benchmark targeting is hugely overparameterized capability.''
In these four cases, different facets of intelligence are being considered: performance, behavior, reasoning, and efficiency.
These conflicting assessments stem from treating these distinct facets under a single term.
If natural language possessed separate terms for $\textrm{intelligence}_X$, where X denoted each facet, debates would become more precise and conflicts easier to disentangle.
This hypothetical refinement illustrates a practical role for conceptual rigor that is often lacking in contemporary AI discourse: clarifying distinctions obscured by ambiguous terminology.

A closely related and equally fraught term is ``understanding''.
Debates concerning whether AI systems genuinely understand are especially important because claims about intelligence are often accompanied by broader questions of meaning and reasoning \cite{MitchellKrakauer2023Understanding}.
As with intelligence, disagreements frequently arise from different underlying assumptions about what constitutes understanding.
Whereas notions of intelligence differ in what features they emphasize, views on understanding often differ along two dimensions: degree and form.

The first dimension, degree, highlights that understanding, like intelligence, exists along a spectrum rather than as an all-or-nothing quality.
However, the nature of this spectrum differs markedly between human and artificial systems.
Human understanding is generally smoothly cumulative, progressing from beginner to expert in a manner aligned with our cognitive processes.
In contrast, AI systems exhibit what has been termed ``jagged intelligence'', where they can be superhuman while failing in the most basic ways \cite{DellAcqua2023JaggedFrontier, Karpathy2024JaggedIntelligence}.
They can solve research-level mathematics problems but nevertheless be mistaken about the most trivial arithmetic statements, such as asserting that ``$9.11 > 9.9$'' \cite{openai_why_9pt11_larger_2024}.
They can speak hundreds of languages, yet are easily confused under simple reversals of word order \cite{berglund2024the}.
As with intelligence, it is unclear whether understanding must be human-like or whether it has another sensible instantiation when applied to artificial systems.

Parallel to questions about the degree of understanding are questions about its form: what kinds of processes, structures, or relations are constitutive of genuine understanding.
We can summarize various views as follows.
Behaviorism attributes understanding primarily to observable behavioral competence rather than to particular internal mechanisms or representations \cite{Ryle1949ConceptMind}.
By contrast, representational and functionalist approaches hold that understanding depends on the organization and causal-functional role of internal states within a cognitive system \cite{Fodor1975,Putnam1967}.
Finally, grounding-based perspectives emphasize that understanding depends not merely on the manipulation of internal representations, but on those representations acquiring semantic significance through appropriate connections to perception, action, or sensorimotor engagement with the world \cite{Harnad1990SymbolGrounding}.

Disagreements about whether current AI systems ``understand'' often reflect differing assumptions about which of the preceding criteria are necessary.
A behaviorist orientation emphasizes successful task performance, while a functionalist perspective asks whether systems instantiate the appropriate internal processes (e.g. whether some form of reasoning or planning is occurring) \cite{Bengio2019FromSystem1to2}.
On the other hand, some representational approaches argue that genuine understanding requires internal world models\footnote{There is ongoing debate about whether current neural networks possess a (partial) world model, particularly for those that play board games \cite{Li2023EmergentWorldRepresentations} or generate videos~\cite{ParkerHolder2025Genie3}.} capable of supporting robust abstraction and causal reasoning \cite{LeCun2022PathAMI, Alter2024LeCunTIME}, although others note that LLMs may already possess some form of meaning through the structure of their internal representations \cite{PiantadosiHill2022Meaning}.
Finally, grounding-based perspectives highlight the need for systems to include sensorimotor modalities \cite{Xu2025Grounding} or more generally, reference to entities outside of language \cite{bender-koller-2020-climbing}.
While these positions are not mutually exclusive, debates about whether current AI systems understand, much like debates about intelligence, are often driven less by empirical disagreement than by differing philosophical assumptions about the criteria understanding requires.

Beyond clarifying individual terms, the field of AI has also evolved through a succession of broader conceptual paradigms.
These paradigms have shaped both research directions and expectations concerning what AI systems can achieve, while also reflecting different relationships between foundational assumptions, scientific understanding, and engineering practice.
Their evolution is examined further in Section \ref{sec:Progress}.

More broadly, conceptual rigor in AI shares affinities with the analytic tradition in philosophy through its emphasis on careful distinctions, explicit meanings, and the dissolution of ambiguities.
Yet the parallel is only partial: although conceptual analysis in the analytic tradition often aims to clarify foundational concepts, terms such as intelligence and understanding in AI do not function in the way that, for example, mass or charge do in physics.\footnote{While there are formal mathematical accounts of intelligence (e.g. \cite{legg2007universal}), they function as idealizations amenable to study rather than as core foundations.}
Their meanings are context-dependent and continually revised in light of new artifacts and capabilities.
Accordingly, notions of intelligence are best understood not as the hard core of a unified research program in the Lakatosian sense, but as interpretive lenses that emphasize different dimensions of performance and competence.
Some of these lenses are tied to specific research agendas while others serve as evaluative standpoints for interpreting the significance of existing systems.
The value of conceptual rigor, then, is not that it delivers conclusive definitions, but that it clarifies the scope and evidential basis of the terms in use.
It helps specify what precisely is being asserted, under what assumptions it holds, and how far a given conclusion can legitimately be extended.
In this way, conceptual rigor does not end disagreements, but it makes them more precise.

Yet conceptual clarification alone cannot determine whether AI systems genuinely reason, generalize, or understand.
Addressing such questions requires not only conceptual analysis, but also empirical and theoretical investigation.
This motivates a second form of rigor: epistemic rigor.

\section{Epistemic Rigor}
\label{sec:EpistemicRigor}

Epistemic rigor provides the standards for generating and validating knowledge.
It includes the use of mathematical proof, the formulation of testable hypotheses, and the careful execution of controlled experiments.
Such methods ground the legitimacy of a scientific discipline.
Without epistemic rigor, it is difficult to determine when results will generalize or why a method succeeds at all.

The notion of epistemic rigor developed here differs from traditional concerns in epistemology and the philosophy of science.
Epistemology has largely focused on the conditions under which beliefs count as knowledge, including questions of justification and skepticism, while philosophy of science has examined the evaluation of scientific theory, including how claims are confirmed, falsified, or revised.
By contrast, the notion of epistemic rigor adopted here is more pragmatic.
Rather than seeking ultimate justification, it emphasizes the procedural and methodological conditions under which scientific results become credible in practice.
In this sense, epistemic rigor is not primarily a theory of knowledge, but a framework for how scientific communities establish reliable understanding in complex domains such as AI, where formal theory and empirical performance are deeply intertwined.

While many earlier areas of AI research achieved a high degree of theoretical grounding\footnote{Earlier approaches were often founded on mathematically grounded methods such as convex optimization, online learning, and probabilistic modeling.}, modern deep learning has shifted the field toward large-scale empirical methods whose successes often outstrip scientific understanding.
In this setting, epistemic rigor is especially important for producing reliable results, given the scale of resources needed to conduct experiments \cite{Cottier2023DollarTrainingCostML, YouOwen2025PowerDemandsFrontierAI} as well as the rapid pace at which results are produced and disseminated \cite{Maslej2023AIIndexReport}.
Yet epistemic rigor concerns more than the validation of isolated empirical findings.
Scientific understanding also requires unifying these findings into a robust and generative body of knowledge.
Accordingly, these aims depend upon three closely related criteria: reproducibility, predictability, and explainability.

\subsection{Reproducibility}

Reproducibility requires that key findings remain robust under ordinary and inevitable variations in experimental conditions, such as differences in infrastructure or implementation.
This allows results to be shared and accepted as established findings.

AI research\footnote{Henceforth, ``AI research'' generally refers to research in modern deep learning, its most influential and relevant subfield.}, when conducted properly, enjoys a comparatively high degree of epistemic rigor with respect to reproducibility.
Experiments are driven by code, data, and models, all of which can be copied and shared.
Furthermore, open source repositories and standardized benchmark datasets enable ideas to spread and results to be quickly verified or disconfirmed.

Reproducibility in AI research, however, remains more fragile than it may initially appear \cite{pmlr-v97-bouthillier19a, henderson2018deep}.
In \cite{pmlr-v97-bouthillier19a}, a useful distinction is made between \textit{results reproducibility} (reimplementation of a method generates statistically similar values) and \textit{inferential reproducibility} (varying experimental setups lead to similar conclusions).
Results reproducibility ensures that outcomes from independent replication of the experiment under the expected sources of variation (e.g. random seed, code implementation) remain stable, while inferential reproducibility ensures that wider claims inferred from the original experimental setup stand up to scrutiny.
Unfortunately, differing implementations of learning algorithms can often lead to significant alterations in performance, thus undermining results reproducibility \cite{henderson2018deep}.
At a broader level, conclusions about which algorithms or architectures perform best are often revised when the experimental settings are changed or expanded, in turn making inferential reproducibility a challenge \cite{lucic2017_are_gans_created_equal, pmlr-v97-bouthillier19a, li2019random}.
Together, such findings illustrate that reproducibility in AI depends not merely on sharing experimental configurations and artifacts, but on careful methodology and transparent reporting of experimental variability.

\subsection{Predictability}

Predictability is central to scientific understanding because it enables researchers to characterize or anticipate system behavior prior to experiment.
Such prediction may proceed deductively through formal mathematical analysis or inductively through the discovery of empirical regularities and laws.

In AI, neural networks have the distinct advantage of being software: they are specified entirely in terms of mathematics and code.
As a result, neural networks and their training dynamics are in principle fully specified and therefore amenable to precise formal analysis.
This contrasts with the situation in the physical sciences, where there is a distinction between the material objects studied by experimentalists and the mathematical models studied by theoreticians.
Consequently, the limits of prediction arise less from misalignment between theory and reality than from the intelligibility of the systems themselves.

In practice, however, the predictive reach of theory is often narrowly circumscribed.
For one, the mathematical objects involved are extremely complicated: neural networks involve many parameters (millions if not trillions), consist of highly nested compositions of nonlinear functions, and are optimized with respect to a nonconvex objective.
Furthermore, in real-world settings, the data involved are complex and noisy, and thus are not amenable to the clean hypotheses used in mathematically tractable settings.
Because precise theoretical prediction is often unavailable, researchers instead rely on reproducibility as a foundation for identifying stable empirical regularities.
In effect, rather than beginning with a theoretical prediction that researchers set out to confirm or falsify, they start with a reproducible baseline and proceed via exploration and discovery.
These contrasting modes of research have been distinguished in \cite{pmlr-v97-bouthillier19a, herrmann2024rethinking_empirical_ml} as \textit{confirmatory} (or \textit{empirical}) research and \textit{exploratory} research, respectively.
The prevalence of the latter over the former means that successful results in AI typically occur as a consequence of experiment rather than through the confirmation of an experimental prediction.

Many specific aspects of deep learning are nevertheless amenable to prediction due to their observed regularity: increasing compute tends to improve performance \cite{kaplan2020scalinglawsneurallanguage}, models retrained on new data catastrophically forget \cite{Kirkpatrick2017catastrophic}, LLMs regurgitate their training data \cite{carlini2021extracting, Nguyen2024UnderstandingTransformers}, and the like.
But such predictions remain limited in scope, typically applying only within closely related settings.
This falls far short of the predictive range of theories in the natural sciences that extend well beyond the domains in which they were developed.
Gravity on the surface of the earth is the same gravity that causes the Moon to orbit the Earth.
Mass spectrometry reveals the chemical composition of ordinary materials and the distant stars.
By comparison, deep learning possesses relatively few predictive theories capable of extrapolating reliably to distant regimes.

Despite these limitations, a small number of predictive frameworks with meaningful extrapolative power have emerged.
The first arises from theories that consider various limits in which the width of a neural network goes to infinity, resulting in greatly simplified dynamics \cite{jacot2018neural, Lee2019WideNeuralNetworksLinear, mei2019meanfield, yang2022tensor}.
This has enabled predictions of how optimal choices of hyperparameters like the learning rate scale with model size \cite{yang2022tensor}.
A second important example is so-called ``scaling laws'' \cite{hestness2017deeplearningscalingpredictable, kaplan2020scalinglawsneurallanguage}.
These laws describe power law behavior in the loss function of a neural network at the end of training as a function of quantities such as compute, data, or model size.
The significance of these laws stems from their empirical validity across many orders of magnitude, allowing neural network behavior to be extrapolated to scales beyond the current frontier.
They thus enable researchers to predict aspects of training outcomes, such as the final loss, before committing resources to large-scale experiments.\footnote{These laws are not without caveats and controversy, and their exact form continues to be revised. See \cite{hooker2025slowdeath} for an overview.}
More broadly, recent work has argued that such developments may represent the early stages of an emerging scientific theory of deep learning centered on the dynamics and statistics of the learning process itself \cite{Simon2026ScientificTheoryDL}.

Nevertheless, AI research remains hindered in its development as an epistemically rigorous science due to limited predictability in key areas.
First, existing theory often fails to specify in advance the conditions under which neural networks will generalize or remain reliable.
In particular, there is no universally robust characterization of when inputs should be regarded as out-of-distribution relative to the training data.
Moreover, the ease with which adversarial examples\footnote{Adversarial examples, in their original form, arise from adding visually imperceptible perturbations to an input image that cause a model to confidently misclassify it.} can be constructed, together with the difficulty of characterizing and preventing them, limits the ability to predict the conditions under which neural networks will behave robustly \cite{GoodfellowShlensSzegedy2015AdversarialExamples}.
Second, both the performance of algorithms and comparisons between them depend critically on the choice of hyperparameters \cite{lucic2017_are_gans_created_equal, henderson2018deep, Bouthillier2021accounting}.
Yet without predictive principles governing how hyperparameters should be selected, practitioners often rely upon extensive search, manual tuning, and expert intuition to obtain strong performance.
This sensitivity makes it difficult to determine whether observed behavior reflects robust properties of an algorithm or artifacts of particular configurations.
Together, these limitations in predicting neural network behavior and algorithmic performance present significant obstacles to epistemic rigor.

\subsection{Explainability}

Scientific explanations seek to account for how and why phenomena occur.
Good explanations contribute to epistemic rigor by enabling generalizable knowledge and counterfactual reasoning.
Ultimately, explanations provide the basis for scientific understanding.

For deep learning, the lack of adequate explanations for its successes has led many to regard the field as ``alchemy'' \cite{Hutson2018AIAlchemy, Arora2019AlchemyControversy, Kolter2024AlchemyNeurIPS}.
This characterization deserves closer inspection, as fields like medicine and neuroscience operate on incomplete theoretical accounts of their underlying mechanisms yet are not labeled as alchemical practices.
What distinguishes deep learning is not simply incomplete understanding, but the nature of its epistemic limitations.
First, it defies effective hierarchical abstraction.
In mature sciences, explanatory questions can often be localized to distinct levels of analysis, where questions are resolved at well-defined scales such as the subatomic or cellular level.
Deep learning provides no such clarity: it is frequently unclear whether observed behavior should be attributed to individual neurons, layer interactions, optimization dynamics, training data, or some combination thereof.
Consequently, the field remains without a systematic framework for decomposing phenomena into distinct components and their associated explanations.
Second, neural networks lack interpretability.\footnote{While explainability and interpretability are often used interchangeably, we regard the latter as providing a foundation for the former. We take interpretability roughly to mean comprehensibility, for example, texture and shape are interpretable features of an image.}
This issue extends beyond mere complexity.
Biological neural networks and large collections of molecules also form complex systems.
The central difficulty is that deep neural networks learn features that are not readily interpretable.
By comparison, traditional machine learning makes use of explicitly defined attributes.
And in other scientific fields, abstractions have components with well-defined features.
Electrons have mass and spin, while molecules have chemical formulae.
However, the feature vectors produced by neural networks live in high-dimensional vector spaces that defy straightforward interpretation.
As such, neural networks are regarded as ``black boxes'' because, while they are structurally transparent, their inner workings elude understanding.

The effectiveness of scaling in deep learning has also reduced the need for explainability.
Traditionally, improving performance meant designing better algorithms, a process requiring significant expertise and understanding.
But with deep learning, approaches that scale effectively often continue to improve with sufficient data and compute.
In turn, continued capability improvements through scaling have become more predictable, while developing good explanations remains a separate and more uncertain endeavor.

Despite these difficulties, deep learning does possess partial explanatory frameworks that contribute meaningfully to epistemic rigor.
Many core components of deep learning are built upon sound theoretical principles arising from classical statistical learning \cite{Vapnik2000, ShalevShwartzBenDavid2014UML}, approximation theory \cite{pinkus1999approximation}, and optimization \cite{nocedal2006numerical, hazan2023introductiononlineconvexoptimization}.
Such theoretical results provide explanatory insight insofar as they necessitate or make plausible what can occur.
This explanatory role is familiar throughout mathematics and statistics.
The central limit theorem, for instance, explains why the normal distribution frequently arises when sampling data.
Similarly, in machine learning, PAC learning bounds \cite{ShalevShwartzBenDavid2014UML}, while loose, help explain why neural networks often generalize better as they are trained on more data.
Likewise, convergence results for gradient-based optimization algorithms in simple settings motivate their application to more complex neural network objectives, while results showing that sufficiently large neural networks can approximate broad classes of functions \cite{pinkus1999approximation} shed light on how neural networks are capable of learning highly complex tasks.
Beyond the core of deep learning, many influential approaches, including reinforcement learning and generative modeling, are likewise grounded in substantial mathematical theory.
Such results contribute to epistemic rigor by providing principled accounts of why particular methods succeed and by clarifying the conditions under which they behave as expected.

Nevertheless, many of the most distinctive features of deep learning remain only partially understood.
In fact, what has made deep learning mysterious and a rich source of surprises is its departure from classical machine learning.
Important examples include the tendency of larger neural networks to generalize better and the success of gradient-based optimization for highly nonconvex loss functions, both of which contradict classical intuitions.
In particular, the classical bias-variance tradeoff suggests that larger, more complex models should generalize more poorly than simpler ones, while simple nonconvex optimization problems suggest that gradient-based methods should become trapped in local minima.
However, ongoing work on phenomena such as double descent \cite{belkin2019reconciling_double_descent}, implicit regularization \cite{soudry2017implicit}, and the effects of overparameterization on the loss landscape \cite{ge2017no, du2018gradient} has provided important insights for why deep learning is able to overcome these anticipated obstacles.
Extending such analyses so that they yield practical guidance in realistic experimental settings would represent a major advance in the explanatory maturity of deep learning.

A further explanatory ambition is the identification of fine-grained causal mechanisms.
This would involve mapping internal neural network components and their interactions to the behaviors they implement, an approach commonly known as mechanistic interpretability.
Ideally, such decompositions would provide functional explanations of how neural networks transform inputs into outputs.
Understanding these internal mechanisms would also enable a range of practical interventions, including updating knowledge in models \cite{wang2024knowledgeediting}, steering models toward desirable behavior \cite{zou2024improving}, and diagnosing failure modes such as hallucinations \cite{xiong2025faithfulretrievalaugmentedgenerationsparse} and coding errors \cite{bui2025correctness}.

Yet, as desirable as explanation and interpretability may be, achieving them remains a significant challenge.
Both are nuanced concepts with competing and sometimes conflicting formulations \cite{LiptonMythos, JainWallace2019Attention, wiegreffe-pinter-2019-attention}.
Moreover, questions about what qualifies as an explanation, including whether explanations must be unique, together with the conceptual difficulty of phenomena such as deception and lying, have drawn attention to the philosophical foundations of interpretability \cite{meloux2025everything, Williams2025MechanisticInterpretabilityNeedsPhilosophy}.
While mathematical approaches to causality could in principle provide a rigorous explanatory framework \cite{pearl2009causality}, their reliance on abstractions such as directed acyclic graphs makes them difficult to apply directly to the low-level mechanisms of neural networks.
In practice, the complexity of deep learning systems leads to substantial underdetermination: multiple plausible explanations may account for the same behavior, many of them proposed post hoc rather than uniquely specified and validated in advance.

Finally, the role of explanation in AI raises deeper questions that remain unresolved.
In traditional scientific domains, explanations are often expected to be both accurate and intelligible, enabling humans to form a coherent understanding of the underlying mechanisms.
However, modern AI systems challenge this expectation.
Neural networks implement highly complex, distributed computations for which no succinct or human-comprehensible description may exist.
If so, it is unclear whether explanation, as traditionally conceived, can remain a central requirement of epistemic rigor, or whether new standards must be developed that relax demands for interpretability.
Moreover, artificial systems are not readily amenable to explanation through theory of mind or high-level behavioral abstractions.
This stands in contrast to human behavior, which is routinely explained in terms of beliefs, desires, and intentions, even in the absence of detailed neural understanding.
The result is a significant explanatory gap between low-level mechanisms and abstract behavioral descriptions of neural networks.
Addressing these issues, including what counts as an explanation and at what level of abstraction, will be essential for developing principled frameworks for interpreting increasingly complex AI systems.\\

Epistemic rigor ultimately concerns the conditions under which empirical findings develop into reliable scientific knowledge.
In AI, this process remains underdeveloped.
While deep learning has achieved remarkable practical successes, its reproducibility, predictability, and explainability each remain limited in important ways.
Strengthening epistemic rigor will therefore be necessary for transforming AI from a discipline driven primarily by empirical performance into one grounded in principled understanding.

\section{Operational Rigor}
\label{sec:OperationalRigor}

Notwithstanding the above challenges, the practical success of AI systems ultimately depends on whether they function reliably and effectively, thereby necessitating operational rigor.
Operational rigor concerns the ability to build and deploy systems that meet defined performance standards or outcome criteria.
It consists of sound methodology: thorough testing and validation, systematic monitoring of performance and failure modes, and the implementation of robust and scalable designs under real-world conditions.
Performance metrics play a central role, since what is measured can be controlled, compared, and improved across deployment contexts.

Operational rigor is not unique to AI.
In established engineering fields such as aeronautics, it ensures that systems behave in accordance with well-understood theoretical constraints.
In other domains, such as medicine, it supports reliable outcomes despite partial understanding of underlying mechanisms.
What distinguishes AI is the role operational rigor plays in enabling the use of systems whose behavior and failure modes are often neither constrained nor predictable from existing theory.
In this setting, operational rigor is not derived from prior understanding; it substitutes for it.
In AI, operational rigor is most prominently realized through two complementary mechanisms: benchmarks, which evaluate performance, and reliability and safety procedures that guide systems toward intended behavior.

\subsection{Benchmarks}

A benchmark represents a standardized suite for performance evaluation that consists of curated datasets, specific tasks, and predefined success metrics.
Quantifying a system's performance through such metrics provides rigor in evaluation that eliminates the ambiguity of human judgment.
This contrasts with earlier evaluation proposals such as the Turing Test \cite{Turing1950Computing}, whose outcomes depend heavily on subjective interpretation.

The value of benchmarks lies in their role as proxies for real-world performance.
Because it is infeasible to evaluate AI systems across every possible input or deployment condition, benchmarks approximate this impossible task through a sufficiently representative set of examples.
Crucially, a benchmark must capture enough of the relevant structure of deployment settings to reliably predict how a system will perform in practice.
Such curation is a form of operational rigor that compensates for limited epistemic guarantees of system behavior outside the benchmark setting.
But benchmarks do more than measure domain-specific generalization; they are often treated as evidence of broader capabilities.
For instance, influential benchmarks such as ImageNet \cite{Deng2009ImageNet} are not primarily valuable because of their specific classification task, since real-world applications extend far beyond the categories considered in such datasets.
Rather, such benchmarks function as epistemic proxies: strong performance is taken as evidence that a model is capable of generalizing across diverse visual environments and tasks.
Operational rigor therefore involves ensuring that such proxies are well-calibrated, meaning that benchmarks track success beyond their original evaluation settings.

Benchmarks also play a central role in operational rigor because they transform performance into an explicit optimization target.
Methods can be directly compared according to standardized metrics, allowing techniques that improve performance to be refined and scaled while less effective ones are discarded.
In this way, benchmark-defined objectives often become achievable through sustained optimization.

Nevertheless, benchmarking is not without its problems.
For strong benchmark performance to be indicative of broader generalization, two conditions must hold: models must not have been trained on benchmark data, and they must learn the intended concepts underlying the benchmark.
The first condition has become increasingly difficult to satisfy in an era in which large AI systems are trained on massive datasets.
Because it is often unclear whether training data substantially overlaps with benchmark data, benchmark scores can be difficult to regard as reliable.
Second, neural networks can fail to learn the underlying task through ``shortcut learning,'' whereby they latch onto spurious correlations.
A well-known study shows that image classifiers are much more likely to recognize cows when the background contains grass rather than a beach \cite{Beery2018RecognitionTerraIncognita, Geirhos2020ShortcutLearning}.
More generally, model performance can degrade sharply under mild perturbations or distribution shifts, showing that high benchmark scores do not necessarily imply robust generalization \cite{hendrycks2018benchmarking, mirzadeh2025gsmsymbolic}.
Another limitation follows from Goodhart's law: once benchmark metrics become optimization targets, they can themselves be exploited.
A well-known example is sycophancy \cite{perez-etal-2023-discovering}, in which chat-based LLMs prioritize agreement with users over truthfulness.
A further limitation is that benchmark rankings are fragile with respect to benchmark selection itself.
Specifically, the ``benchmark lottery'' hypothesis \cite{dehghani2021benchmarklottery} argues that small changes in the choice of benchmark tasks can substantially alter the relative ranking of methods, implying that perceived algorithmic superiority may be an artifact of how benchmarks are constructed.

As a result, designing effective and meaningful benchmarks remains one of the most consequential challenges in AI research.
Expanding the number and diversity of benchmarks is necessary for broadening the range of capabilities that can be evaluated, while ensuring that benchmarks faithfully capture real-world performance remains an enduring challenge.
The latter becomes especially difficult as desired tasks grow increasingly complex and open-ended.
For instance, current limitations such as the inability of AI systems to perform long-horizon continual learning\footnote{Continual learning involves adapting to new data over time without requiring expensive retraining.} may persist in part due to the absence of strong benchmarks for evaluating this capability.
In this regard, benchmark design cannot be separated from the other forms of rigor discussed previously: conceptual rigor is required to determine how capabilities should be defined and scoped, while epistemic rigor is needed to understand existing failure modes and whether proposed benchmarks are appropriate for addressing them.

\subsection{Reliability and Safety}

Operational rigor is needed to ensure that AI systems are reliable and adhere to principles such as helpfulness, honesty, and safety.
Indeed, without explicit efforts to instill such properties, they can easily behave in harmful or undesirable ways \cite{gehman2020realtoxicityprompts, elatillah2023ai}.
In developing AI to become a widespread technology, researchers and engineers have succeeded in making these systems more controllable and useful in a variety of ways:

One broad class of methods augments LLMs with additional tools or computation \cite{Schick2023Toolformer}.
When LLMs can invoke external tools, they are no longer forced to solve every problem internally; instead, they can delegate subtasks to specialized systems.
In doing so, LLMs inherit the reliability of their suite of tools, transferring the burden of correctness to proper tool use.
Modern LLMs are thus no longer isolated chatbots, but now function as natural-language interfaces to complex systems.
Operational rigor becomes less a matter of the performance of an isolated model and more a matter of system orchestration: deciding which component acts and when.

Operational rigor also includes eliciting latent capabilities already present within models.
Techniques such as prompt engineering, extended computation, and self-critique can substantially improve reliability and task performance without modifying model parameters \cite{wei2022chainofthought, madaan2023selfrefine}.
In addition, detailed system prompts are often prepended during deployment to provide persistent behavioral guidance across user interactions.
Although such methods do not provide guarantees of correctness, they offer a measure of control over behavior despite limited understanding of how models function.

A complementary approach involves directly shaping model behavior through post-training.
After the initial pre-training stage in which LLMs are trained to predict the next token, they undergo additional post-training via instruction fine-tuning and reinforcement learning from human feedback \cite{wei2022finetuned, Christiano2017RLHF, Ouyang2022TrainingLMsToFollowInstructions}.
These later stages refine pretrained models into systems capable of reliably following user requests while reinforcing behaviors preferred by human evaluators.
Here, operational rigor depends heavily on careful dataset and reward design: diverse task distributions and finely specified human preference criteria shape how models balance competing objectives such as harmlessness and deference to user intent.

Yet, despite ongoing efforts to strengthen current AI systems, they still suffer from many weaknesses.
LLMs often hallucinate and fail to follow instructions properly, sometimes quite spectacularly \cite{BBCNews2024_GoogleAISearchErrors, Barr2025_ReplitAI_CatastrophicDeletion}.
Their vulnerability to adversarial attacks and jailbreaks makes them susceptible to misuse \cite{xu-etal-2024-comprehensive, xu2025surveyattackslargelanguage, attacks_LVLMs}.
Researchers have also shown that training data can be easily poisoned, making it possible to embed malicious behavior inside models \cite{ramirez2022poisoning,zhao2025datapoisoning}.
Furthermore, open-weight models make the situation even more fragile, since nothing inherently prevents users from modifying these models to override preexisting safety mechanisms.
These limitations reveal a central difficulty for operational rigor: current methods for controlling model behavior have not scaled at the same pace as model capability itself.
As a result, ensuring robust and reliable behavior under realistic conditions remains an unresolved challenge.

One potential solution is the use of formal verification methods, which aim to provide mathematical guarantees that systems satisfy specified constraints, thereby reducing reliance on empirical evaluation alone.
Existing work includes certifying robustness to bounded perturbations, proving safety conditions for neural-network-controlled systems, and validating model-generated outputs using formal verifiers or proof assistants \cite{albarghouthi2021introduction, xiang2018verification, ivanov2019verisig, Hubert2025AlphaProof}.
However, these methods remain difficult to scale to modern architectures, or else are generally limited to settings in which correctness conditions can be precisely specified.\\

Operational rigor in AI departs in important respects from traditional philosophical accounts of scientific rigor.
In much of the philosophy of science, empirical evaluation primarily serves epistemic aims such as establishing causal relationships, supporting explanations, or refining theories.
In contemporary AI, however, evaluation increasingly functions not only as a means of assessing systems, but also as a primary mechanism for improving them.
This does not eliminate epistemic concerns, but it changes the relationship between performance and scientific understanding in ways that distinguish AI from many other fields of science and engineering.

\section{Rigor and Progress}
\label{sec:Progress}

The trajectory of scientific and technological progress depends not only on what modes of rigor are present within a field, but also on how they interact and develop over time.
In physics, advances such as spaceflight and semiconductors emerged from conceptual frameworks capable of supporting strong prediction and reliable engineering.
In biology, progress has depended on connecting explanations across molecular, cellular, and evolutionary scales, allowing phenomena such as heredity and natural selection to be understood even when living systems remain difficult to predict and control.

In AI, by contrast, the relationship between rigor and progress has been striking.
AI has advanced rapidly as a technology despite limited conceptual clarity or theoretical grounding.
This development may be understood through the shifting roles played by conceptual, epistemic, and operational rigor.
We now examine how these roles have shaped AI across paradigms and patterns of progress, as well as how they may continue to shape the field’s future direction.

\subsection{The Distinctive Structure of AI Progress}

Progress in modern AI has not emerged from the uniform development of the three forms of rigor.
Instead, operational rigor has come to assume a dominant role in guiding progress while conceptual clarification and scientific understanding have often lagged behind.
This asymmetry reflects deeper structural features of the field itself.

One such feature is that AI exhibits a tight feedback loop between evaluation and development: the same metrics used to assess progress are themselves targets for further optimization.
Operational rigor therefore functions not merely as a way of evaluating systems after they have been built, but also as a mechanism through which they are improved.
This configuration has no clear analogue in many other fields.
In medicine, practical and ethical constraints prevent benchmark-driven iteration, while in physics and traditional engineering, manual design guided by theory drives improvement.
By contrast, AI systems are trained to optimize metrics directly, enabling increasingly strong performance even in the absence of a mature scientific theory.

A second structural feature is that AI produces the very artifacts it studies.
Models are not merely instruments for investigating intelligence; they are themselves often regarded as new instances of intelligent behavior.
Each generation of systems introduces new capabilities, limitations, and failure modes, thereby expanding the domain that conceptual and epistemic inquiry must explain.
This differs from the natural sciences, where the principal objects of inquiry, such as atoms, organisms, and planets, exist largely independently of the field itself.
In AI, however, the objects of study are endogenous to the field’s own methods of development.

Together, these structural features help explain why progress in AI can be both rapid and uneven.
Benchmark optimization makes it possible to develop methods that generate measurable improvements without first possessing a substantive theory of the systems being improved.
At the same time, the continual production of new artifacts ensures that conceptual clarification and explanatory frameworks are always responding to a moving target.
The result is a form of progress in which capabilities advance faster than the conceptual and scientific frameworks needed to fully understand them.

\subsection{Rigor Across Paradigms}

The predominance of operational over epistemic rigor described above is associated with the current deep learning paradigm rather than with AI as a whole.
Throughout the history of AI, different paradigms have embodied varying balances between the forms of rigor, reflecting changing assumptions about how intelligence should be modeled and realized.
The current imbalance between operational and epistemic rigor is therefore historically contingent rather than inevitable, and may not persist in future paradigms.

This variability can already be seen in earlier approaches to AI.
The symbolic approach represented a paradigm that viewed intelligence as the manipulation of explicit representations governed by formal rules \cite{russell_norvig_2020}.
This paradigm possessed strong epistemic rigor because systems were interpretable and their operations could be traced through with precision.
Yet these systems performed poorly outside carefully constrained environments.
Their brittleness and inability to scale limited their effectiveness, and their restricted domains of application left little scope for the kind of large-scale benchmark optimization characteristic of AI today.

Learning-based paradigms relaxed the requirement that intelligent behavior be specified explicitly in advance.
Instead, systems learn statistical structure from data.
This paradigm developed along two distinct but related lines: classical statistical learning and connectionism \cite{Vapnik2000, RumelhartMcClelland1986PDP1}.
The classical statistical learning paradigm developed a strong degree of epistemic rigor: progress was closely tied to developing mathematically tractable models, including linear classifiers, kernel methods, and decision trees, alongside the development of statistical learning theory providing generalization guarantees \cite{ShalevShwartzBenDavid2014UML}.
Because model classes remained comparatively tractable, theoretical analysis continued to play a central role in guiding progress.
Moreover, simply scaling model size did not reliably produce improved generalization in this setting, as increasingly complex models were prone to overfitting.

By contrast, the connectionist paradigm departed significantly.
This approach adopted a comparatively epistemically modest stance: it relied less on manually specified structure and more on learning procedures capable of discovering useful representations directly.
This relaxation of epistemic constraints would prove extraordinarily consequential. While early connectionist systems achieved mixed success, the paradigm eventually gave rise to deep learning, whose large-scale neural networks proved capable of dramatic empirical improvements.
Crucial to the success of this approach were technological developments that enabled optimization to scale effectively, particularly advances in computational hardware and the availability of large datasets.
As a consequence, deep learning systems achieved unprecedented performance despite limited theoretical understanding of why they worked so well.

These foundational paradigms, consisting of symbolic AI, classical statistical learning, and deep learning, illustrate the dynamic relationship between our three forms of rigor.
Conceptual rigor shapes the paradigms through which intelligence is modeled and realized, influencing how epistemic and operational rigor interact within a given approach.
More recent developments suggest how this interplay may continue to evolve.
One example is the growing incorporation of neuro-symbolic elements into modern AI systems, which attempt to recover some degree of predictability and explainability by combining learned systems with external tools, thus partially restoring epistemic rigor.
At the same time, the rise of agentic systems places increasing emphasis on long-horizon planning, autonomous action, and open-ended interaction.
Such capabilities increase performance demands while simultaneously introducing new epistemic difficulties: the opacity of neural network internals is now compounded by the complexity of agentic behavior.
Operational rigor is likely to become even more dominant relative to epistemic rigor in this agentic setting, absent major theoretical breakthroughs.
Yet the evolution of AI suggests that future paradigms may again alter the balance between operational and epistemic rigor.

\subsection{Future Progress}

Within the current deep learning paradigm, future progress is likely to depend increasingly on strengthening areas of rigor that remain comparatively underdeveloped.
Two prominent examples are the pursuit of increasingly general intelligence and the problem of alignment, each of which places different demands on rigor.
Progress toward general intelligence currently favors operational rigor, whereas alignment depends more heavily on conceptual and epistemic rigor.

Artificial general intelligence (AGI) has long served as an aspirational goal: a system capable of performing the full range of cognitive tasks associated with human intelligence.
Yet, like intelligence itself, the term AGI remains overloaded.
It has been associated with ideas ranging from economic and social upheaval \cite{korinek2024_economic_policy_challenges_age_ai} to eschatological singularity narratives \cite{kurzweil2024_singularity_is_nearer} and science-fiction scenarios \cite{CaveDihal2019Hopes}.
The wide range of meanings associated with AGI\footnote{Some researchers doubt whether AGI names a coherent concept at all \cite{Gopnik2020CritiqueAGI,InformationBottleneck2025LeCun,Amodei2025_AnthropicInterview}.} is not problematic per se, since many important concepts resist precise definition.
But AGI, beyond being a research program, is increasingly discussed in public and policy contexts \cite{AguerayArcasNorvig2023AGI,korinek2024_economic_policy_challenges_age_ai}, including by CEOs of major industrial AI labs, and therefore merits clarification whenever it is invoked.

The conceptual ambiguity of AGI implies that a critical bottleneck for progress lies in operationalizing the definition of AGI.
Such an operational definition supports both conceptual and operational rigor because it clarifies the concept of AGI by grounding it in performance measures that can be objectively assessed.
Recent proposals include frameworks for defining and evaluating AGI \cite{hendrycks2025definitionagi, burnell2026measuring}, as well as taxonomic approaches such as \cite{Morris2024LevelsAGI}.
The creation of benchmarks grounded in operational definitions of AGI may enable progress to occur even when epistemic rigor remains underdeveloped and a comprehensive scientific theory of intelligence is lacking.
This possibility is reflected in a prominent perspective within contemporary AI research: that scaling compute may be largely sufficient for increasingly general capabilities to emerge \cite{altman2025_three_observations}.
Closely related views argue that intelligence itself may largely be captured through optimization against suitably defined reward signals \cite{silver2021reward}.
In such cases, developing AGI may depend less on possessing a rigorous scientific theory than on arranging the conditions for it to arise.
Operational rigor therefore emerges as a central mechanism through which AGI progress may occur.

The case of alignment, however, provides a striking reversal of the priority of operational rigor that we have seen thus far.
Here, alignment refers to the design of systems that are aligned with the values of their designers or, from a humanist perspective, that serve and benefit humanity \cite{Russell2019HumanCompatible, Gabriel2020ValuesAlignment}.
In the limit of highly capable agentic systems, alignment depends critically on conceptual and epistemic rigor.
Unlike intelligence and AGI, which concern what systems can do, alignment additionally concerns what they must not do.
As a result, determining which behaviors must be avoided depends on conceptual rigor for selecting an appropriate normative framework.
In this regard, the trolley dilemmas of moral philosophy may become matters of practical concern when aligning AI systems.

Preventing powerful AI from directly causing or enabling catastrophic harm (e.g. designing biological weapons or destroying critical infrastructure) requires more than benchmark performance \cite{shah2025approachtechnicalagisafety}.
A satisfactory solution involves understanding whether AI systems adopt intended objectives beyond their training conditions \cite{langosco2022_goal_misgeneralization_rl, hubinger2019risks} and, if not, developing principled accounts of the resulting failure modes.
Moreover, alignment based primarily on benchmark performance is insufficient in a world where increasingly capable systems become widely accessible.
As the cost of developing powerful models declines, malicious actors may optimize systems toward harmful objectives just as readily as benign actors optimize them toward beneficial ones.
For this reason, alignment ultimately requires more than producing systems that behave appropriately in isolated settings.
It requires understanding AI at a level that allows misaligned objectives to be anticipated and mitigated.

\section{Conclusion}

We examined rigor in AI through the differing roles it plays, focusing on the ambiguity, uncertainty, and tensions that arise within the field.
Our contribution was to divide such problems along conceptual, epistemic, and operational dimensions, and to clarify the role that rigor plays in each.
While a scientific or engineering approach to these issues would focus on their solutions, our philosophical analysis mapped the overarching structure, interrelationships, and broader forms their solutions might take.
This bird’s-eye view is valuable for understanding a multidisciplinary field such as AI because the latter is more than just the union of its constituent problems; it also includes the interactions among them and how they co-evolve.

In this regard, the framework developed here helps explain why modern AI has achieved technological capabilities that outpace scientific understanding, while situating this pattern within the broader evolution of conceptual paradigms.
Moreover, our analysis not only identified promising avenues for future progress, but also provides a basis for reasoning about what AI as a mature field might look like, since rigor is a prerequisite for maturity.
While the exact trajectory is difficult to predict, our three-part framework offers insights into several possible requirements:

Conceptual rigor demands that critical concepts shaping the design and evaluation of systems, such as AGI and alignment, be articulated in ways that enable reliable assessment and coordination.
As AI diffuses throughout society and assumes greater geopolitical significance, a more refined terminology for setting expectations and calibrating the capabilities of such systems will become essential.
Epistemic rigor entails that a well-developed science of AI would exhibit far greater predictability and explainability than it does currently.
A more comprehensive body of laws and theoretical results would tightly constrain experimental design; advances in model interpretability would enable actionable interventions across a wide range of systems; and causal explanations would integrate relevant algorithmic and architectural components at the appropriate scales.
Operational rigor, in turn, would ideally yield systems so reliable and robust that many of today’s failure modes, such as hallucinations and jailbreaks, would become sufficiently rare that they are no longer expected to occur.
Additionally, it would enable a technological landscape in which defensive capabilities consistently exceed those of potential adversaries, rendering attacks either infeasible or limited in their capacity to cause harm.

Meeting such challenges will require advancements across many disciplinary boundaries, each demanding the appropriate role for rigor.
Developing and integrating the forms of rigor examined here will be necessary, though far from straightforward.
Yet doing so will be essential for AI to become a mature science and reliable technology.

\bibliographystyle{unsrt}
\bibliography{references}

\end{document}